\documentclass[sigconf]{acmart}
\AtBeginDocument{%
  \providecommand\BibTeX{{%
    \normalfont B\kern-0.5em{\scshape i\kern-0.25em b}\kern-0.8em\TeX}}}

\usepackage{pifont}
\usepackage{multirow}
\usepackage{url}

\usepackage{amsmath,amssymb,amsfonts, bm}
\usepackage{colortbl} 
\usepackage{xcolor}
\usepackage{array}  
\usepackage{subfigure}

\usepackage[ruled,linesnumbered]{algorithm2e}
\usepackage[export]{adjustbox}
\usepackage{multicol}
\usepackage{fancyhdr}
\SetKwRepeat{Do}{do}{while}
\usepackage{setspace}
\usepackage{hyperref}

\newtheorem{theorem}{Theorem}
\newtheorem{assumption}{Assumption}
\newtheorem{definition}{Definition}
\newtheorem{lemma}{Lemma}

\hypersetup{
    colorlinks=true,
    linkcolor=blue,
    filecolor=blue,      
    urlcolor=blue,
    citecolor=cyan,
}

\copyrightyear{2023}
\acmYear{2023}
\setcopyright{acmlicensed}\acmConference[WWW '23]{Proceedings of the ACM
Web Conference 2023}{May 1--5, 2023}{Austin, TX, USA}
\acmBooktitle{Proceedings of the ACM Web Conference 2023 (WWW '23), May
1--5, 2023, Austin, TX, USA}
\acmPrice{15.00}
\acmDOI{10.1145/3543507.3583426}
\acmISBN{978-1-4503-9416-1/23/04}

\begin{document}
\title{To Store or Not? Online Data Selection for Federated Learning with Limited Storage}

\author{Chen Gong}
\email{gongchen@sjtu.edu.cn}
\affiliation{
    \institution{Shanghai Jiao Tong University}
    \city{Shanghai}
    \country{China}
}
\author{Zhenzhe Zheng}
\email{zhengzhenzhe@sjtu.edu.cn	}
\affiliation{
    \institution{Shanghai Jiao Tong University}
    \city{Shanghai}
    \country{China}
}
\author{Yunfeng Shao}
\email{shaoyunfeng@huawei.com}
\affiliation{
    \institution{Huawei Noah’s Ark Lab}
    \city{Beijing}
    \country{China}
}
\author{Bingshuai Li}
\email{libingshuai@huawei.com}
\affiliation{
    \institution{Huawei Noah’s Ark Lab}
    \city{Beijing}
    \country{China}
}
\author{Fan Wu}
\email{fwu@cs.sjtu.edu.cn}
\affiliation{
    \institution{Shanghai Jiao Tong University}
    \city{Shanghai}
    \country{China}
}
\author{Guihai Chen}
\email{gchen@cs.sjtu.edu.cn}
\affiliation{
    \institution{Shanghai Jiao Tong University}
    \city{Shanghai}
    \country{China}
}

\begin{abstract}
Machine learning models have been deployed in mobile networks to deal with massive data from different layers to enable automated network management and intelligence on devices. 
To overcome high communication cost and severe privacy concerns of centralized machine learning, federated learning (FL) has been proposed to achieve distributed machine learning among networked devices. 
While the computation and communication limitation has    been widely studied, the impact of on-device storage on the performance of FL is still not explored. Without an effective data selection policy to filter the massive streaming data on devices, classical FL can suffer from much longer model training time ($4\times$) and significant inference accuracy reduction ($7\%$), observed in our experiments. 
In this work, we take the first step to consider the online data selection for FL with limited on-device storage. We first define a new data valuation metric for data evaluation and selection in FL with theoretical guarantees for speeding up model convergence and enhancing final model accuracy, simultaneously. 
We further design {\ttfamily ODE}, a framework of \textbf{O}nline \textbf{D}ata s\textbf{E}lection for FL, to coordinate networked devices to store valuable data samples. 
Experimental results on one industrial dataset and three public datasets show the remarkable advantages of {\ttfamily ODE} over the state-of-the-art approaches. Particularly, on the industrial dataset, {\ttfamily ODE} achieves as high as $2.5\times$ speedup of training time and $6\%$ increase in inference accuracy, and is robust to various factors in  practical environments. 
\end{abstract}



\settopmatter{printfolios=true}
\maketitle
\section{Introduction}
\label{section: introduction}
The next-generation mobile computing systems require effective and efficient management of mobile networks and devices in various aspects, including resource provisioning~ \cite{giordani2020toward, bega2019deepcog}, security and intrusion detection~\cite{bhuyan2013network}, quality of service guarantee~\cite{cruz1995quality}, and performance monitoring~\cite{kolahi2011performance}. 
Analyzing and controlling such an increasingly complex mobile network with traditional human-in-the-loop approaches~\cite{nunes2015survey} will not be possible, due to  low-latency requirement~\cite{siddiqi20195g}, massive real-time data and complicated correlation among   data~\cite{aggarwal2021machine}.
For example, in network traffic analysis, a fundamental task in mobile networks, routers  can receive/send as many as $5000$ packets ($\approx5$MB) per second. It is impractical to manually analyze such a huge quantity of high-dimensional data within milliseconds.
Thus, machine learning models have been widely applied to discover pattern behind high-dimensional networked data, enable data-driven network control, and fully automate the mobile network operation~\cite{banchs2021network,rafique2018machine, ayoubi2018machine}.

Despite that ML model overcomes the limitations of human-in-the-loop approaches, its good performance highly relies on the huge amount of high quality data for model training~\cite{deng2019deep}, which is hard to obtain in mobile networks as the data is resided on heterogeneous devices in a distributed manner. 
On the one hand, an on-device ML model trained locally with limited data and computational resources is unlikely to achieve desirable inference accuracy and generalization ability~\cite{zeng2021mercury}. 
On the other hand, directly transmitting data from distributed networked devices to a cloud server for centralized learning (CL) will bring prohibitively high communication cost and severe privacy concerns~\cite{ li2021hermes, mireshghallah2021not}. 
Recently, federated learning (FL)~\cite{mcmahan2017communication} emerges as a distributed privacy-preserving ML paradigm to resolve the above concerns, which allows networked devices to upload local model updates instead of raw data and a central server to aggregate these local models into a global model.

\textbf{Motivation and New Problem.}
For applying FL to mobile networks, we identify two unique properties of networked devices: \emph{limited on-device storage} and \emph{streaming networked data}, which have not been fully considered in FL literature. 
(1) \emph{Limited on-device storage}: 
due to the hardware constraints, mobile devices have restricted storage volume for each mobile application and service, and can reserve only a small space to store training data samples for ML without compromising the quality of other services.  
For example, most smart home routers have only $9$-$32$MB storage~\cite{url_buffer} 
and thus only tens of training data samples can be stored. 
(2) \emph{Streaming networked data}: data samples are continuously generated/received by mobile devices in a streaming manner, and we need to make online decisions on whether to store each generated data sample. 

Without a carefully designed data selection policy to maintain the data samples in storage, the empirical distribution of stored data could deviate from the true data distribution and also contain low-quality data, which further complicates the notorious problem of not independent and identically distributed (Non-IID) data distribution in FL~\cite{karimireddy2020scaffold, li2020federated}. 
Specifically, the naive random selection policy significantly degrades the performance of classic FL algorithms in both model training and inference, with more than $4\times$ longer training time and $7\%$ accuracy reduction, observed in our experiments with an industrial network traffic classification dataset shown in Figure \ref{fig: motivation} (detailed discussion is shown in Appendix~\ref{appendix: motivating experiments}).
This is unacceptable in modern mobile networks, because the longer training time reduces the timeliness and effectiveness of ML models in dynamic environments, and accuracy reduction results in failure to guarantee the quality of service~\cite{cruz1995quality} and incurs extra operational expenses~\cite{akbari2021look} as well as security breaches~\cite{tahaei2020rise, arora2016big}. 
Therefore, a fundamental problem when applying FL to mobile network is \emph{how to filter valuable data samples from on-device streaming data to simultaneously accelerate training convergence and enhance inference accuracy of the final global model?}
\begin{figure}
    \vspace{-0.1cm}  
    \setlength{\belowcaptionskip}{-0.3cm}   
    \subfigure[Convergence Time]{
    \includegraphics[width=0.2\textwidth]{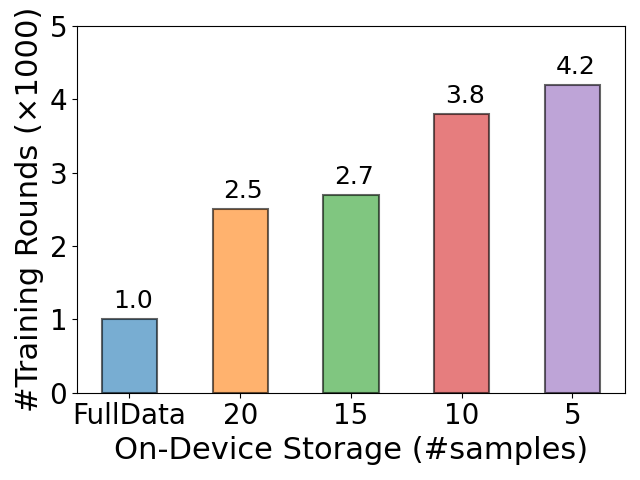}
    }
    \subfigure[Inference Accuracy]{
    \includegraphics[width=0.2\textwidth]{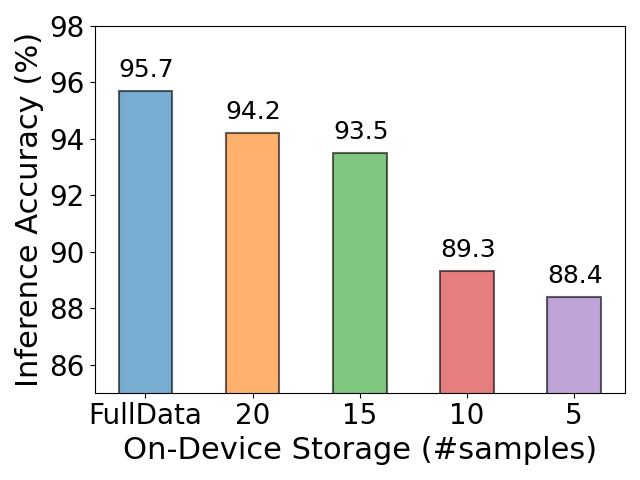}
    }
    \vspace{-0.3cm}
    \caption{To investigate the impact of on-device storage on FL model training, we conduct experiments on an industrial traffic classification dataset with $30$ mobile devices and $35,000+$ data samples under different storage capacities.}
    \label{fig: motivation}
    \vspace{-0.2cm}
\end{figure}

\textbf{Design Challenges.}
The design of such an online data selection framework for FL involves three key challenges:

\noindent\textit{(1) There is still no theoretical understanding about the impact of local on-device data on the training speedup and accuracy enhancement of global model in FL.}
Lacking information about raw data and local models of the other devices, it is challenging for one device to figure out the impact of its local data sample on the performance of the global model.
Furthermore, the sample-level correlation between convergence rate and model accuracy is still not explored in FL, and it is non-trivial to simultaneously improve these two aspects through one unified data valuation metric.

\noindent\textit{(2) The lack of temporal and spatial information complicates the online data selection in FL.}
For streaming data, we could not access the data samples coming from the future or discarded in the past. 
Lacking such \textit{temporal information}, one device is not able to leverage the complete statistical information (\textit{e.g.,} unbiased local data distribution) for accurate data quality evaluation
, such as outliers and noise detection~\cite{DBLP:conf/mobisys/ShinLLL22, li2021sample}. 
Additionally, due to the distributed paradigm in FL, one device cannot conduct effective data selection without the knowledge of other devices' stored data and local models, which can be called as \textit{spatial information}. This is because the valuable data samples selected locally could be overlapped with each other and the local valuable data may not be the global valuable one.

\noindent\textit{(3) The on-device data selection needs to be low computation-and-memory-cost due to the conflict of limited hardware resources and requirement on quality of user experience.}
As the additional time delay and memory costs introduced by online data selection process would degrade the performance of mobile network and user experience, the real-time data samples must be evaluated in a computation and memory efficient way. 
However, 
increasingly complex ML models lead to high computation complexity as well as large memory footprint for storing intermediate model outputs during the data selection process.

\textbf{Limitations of Related Works.}
The prior works on data evaluation and selection in ML failed to solve the above challenges.\\
(1) The data selection methods in CL, such as leave-one-out test~\cite{cook1977detection}, Data Shapley~\cite{ghorbani2019data} and Importance Sampling~\cite{mireshghallah2021not,  zhao2015stochastic}, are not appropriate for FL due to the first challenge: they could only measure the value of each data sample corresponding to the local model training process, instead of the global model in FL. \\
(2) The prior works on data selection in FL did not consider the two new properties of FL devices.
Mercury~\cite{zeng2021mercury}, FedBalancer~\cite{DBLP:conf/mobisys/ShinLLL22} and the work from Li \textit{et al.}~\cite{li2021sample} adopted importance sampling framework~\cite{zhao2015stochastic} to select the data samples with high loss or gradient norm 
but failed to solve the second challenge: these methods all need to inspect the whole dataset for normalized sampling weight computation 
as well as noise and outliers removal~\cite{hu2021does, li2021sample}.

\textbf{Our Solutions.}
To solve the above challenges, we design {\ttfamily ODE}, an online data selection framework that coordinates networked devices to select and store valuable data samples locally and collaboratively in FL, with theoretical guarantees for accelerating model convergence and enhancing inference accuracy, simultaneously.

In {\ttfamily ODE}, we first theoretically analyze the impact of an individual local data sample on the convergence rate and final accuracy of the global model in FL. We discover a common dominant term in these two analytical expressions, 
which can thus be regarded as a reasonable data selection metric in FL.
Second, considering the lack of temporal and spatial information, we propose an efficient method for clients to approximate this data selection metric by maintaining a local gradient estimator on each device and a global one on the server. 
Third, to overcome the potential overlap of the stored data caused by distributed data selection, we further propose a strategy for the server to coordinate each device to store high valuable data from different data distribution regions. 
Finally, to achieve the computation and memory efficiency, we propose a simplified version of {\ttfamily ODE}, which replaces the full model gradient with partial model gradient to concurrently reduce the computation and memory costs of the data evaluation process. 

\textbf{System Implementation and Experimental Results.}
We evaluated {\ttfamily ODE} on \textit{three public} tasks: synthetic task (ST)~\cite{caldas2018leaf}, Image Classification (IC)~\cite{xiao2017/online} and Human Activity Recognition (HAR)~\cite{ouyang2021clusterfl,li2021meta}, as well as \textit{one industrial} mobile traffic classification dataset (TC) collected from our $30$-days deployment on $30$ ONTs in practice, consisting of $560,000+$ packets from $250$ mobile applications. 
We compare {\ttfamily ODE} against three categories of data selection baselines: random sampling~\cite{vitter1985random}, data selection for CL~\cite{DBLP:conf/mobisys/ShinLLL22, li2021sample, zeng2021mercury} and data selection for FL~\cite{DBLP:conf/mobisys/ShinLLL22,li2021sample}.
The experimental results show that {\ttfamily ODE} outperforms all these baselines, achieving as high as $9.52\times$ speedup of model training and $7.56\%$ increase in final model accuracy on ST, $1.35\times$ and $1.4\%$ on IC, $2.22\times$ and $6.38\%$ on HAR, $2.5\times$ and $6\%$  on TC, with low extra time delay and memory costs. 
We also conduct detailed experiments to analyze the robustness of {\ttfamily ODE} to various environment factors and its component-wise effect.

\textbf{Summary of Contributions.} 
(1) To the best of our knowledge, we are the first to identify two new properties of applying FL in mobile networks: \textit{limited on-device storage} and \textit{streaming networked data}, and demonstrate its enormity on the effectiveness and efficiency of model training in FL. 
(2) We provide analytical formulas on the impact of an individual local data sample on the convergence rate and the final inference accuracy of the global model, based on which we propose a new data valuation metric for data selection in FL with theoretical guarantees for accelerating model convergence and improving inference accuracy, simultaneously. 
Further, we propose {\ttfamily ODE}, an online data selection framework for FL, to realize the on-device data selection and cross-device collaborative data storage. 
(3) We conduct extensive experiments on three public datasets and one industrial traffic classification dataset to demonstrate the remarkable advantages of {\ttfamily ODE} against existing methods.

\section{Preliminaries}
\label{Preliminaries}
In this section, we present the learning model and training process of FL.
We consider the synchronous FL framework~\cite{mcmahan2017communication, li2018federated, karimireddy2020scaffold}, where a server coordinates a set of mobile devices/clients 
$C$ to conduct distributed model training. 
Each client $c\!\in\!C$ generates data samples in a streaming manner with a velocity $v_c$. We use $P_c$ to denote the client $c$'s underlying distribution of her local  data, and $\tilde{P}_c$ to represent the empirical distribution of the data samples $B_c$ stored in her local storage. 
The goal of FL is to train a global model $w$ from the locally stored data $\tilde{P}\!=\!\bigcup_{c\in C} \tilde{P}_c$ with good performance with respect to the underlying unbiased data distribution ${P}\!=\!\bigcup_{c \in C} P_c$:
\begin{equation}
    \begin{small}
    \min\limits_{w \in \mathbb{R}^n} F(w)=\sum\limits_{c \in C}\zeta_c \cdot  F_c(w),
    \end{small}
    \label{equation: global objective}
    \nonumber
\end{equation}
where $\zeta_c=\frac{v_c}{\sum_{c'\!\in\!C}v_{c'}}$ denotes the normalized weight of each client, $n$ is the dimension of model parameters, $F_c(w)=\mathbb{E}_{(x,y)\sim P_c}[l(w,x,y)]$ is the expected loss of the model $w$ over the true data distribution of client $c$. We also use $\tilde{F}_c(w)=\frac{1}{|B_c|}\sum_{x,y\in B_c}l(w,x,y)$ to denote the empirical loss of model over the data samples stored by client $c$.

In this work, we investigate the impacts of each client's limited storage on FL, and consider the widely adopted algorithm Fed-Avg~\cite{mcmahan2017communication} for easy illustration\footnote{Our results for limited on-device storage can be extended to other FL algorithms, such as FedBoost\cite{hamer2020fedboost}, FedNova\cite{wang2020tackling}, FedProx\cite{li2018federated}.}. 
Under the synchronous FL framework, the global model is trained by repeating the following two steps for each communication round $t$ from 1 to $T$:

\noindent\textbf{(1) Local Training:} In the round $t$, the server selects a client subset $C_t\!\subseteq\! C$ to participate in the training process. Each participating client $c\!\in\! C_t$ downloads the current global model $w_{\mathrm{fed}}^{t-1}$ (the ending global model in the last round), and performs model updates with the locally stored data for $m$ epochs:
\begin{equation}
    \begin{small}
    w_{c}^{t,i}\gets w_c^{t,i-1}-\eta\nabla_w\tilde{F}_c(w_c^{t,i-1}),\ i=1,\cdots,m
   \end{small}
   \label{eq: FL model update}
\end{equation}
where the starting local model $w_c^{t,0}$ is initialized as $w_{\mathrm{fed}}^{t-1}$, and $\eta$ denotes the learning rate.

\noindent\textbf{(2) Model Aggregation:} Each participant client $c\!\in\! C_t$ uploads the updated local model $w_c^{t,m}$, and the server aggregates them to generate a new global model $w_{\mathrm{fed}}^t$ by taking a weighted average:
\begin{equation}
    \begin{small}
    w_{\mathrm{fed}}^t\gets\sum\limits_{c\in C_t}\zeta_c^t \cdot w_{c}^{t,m},
    \end{small}
\label{eq: FL model aggregation}
\end{equation}
where $\zeta_c^t=\frac{v_c}{\sum_{c'\in C_t}v_{c'}}$ is the normalized weight of client $c$.

In the scenario of FL with limited on-device storage and streaming data, we have an additional data selection step for clients:

\noindent\textbf{(3) Data Selection:} In each round $t$, once receiving a new data sample, the client has to make an online decision on whether to store the new sample (in place of an old one if the storage area is fully occupied) or discard it. The goal of this data selection process is to select valuable data samples from streaming data for model training in the coming rounds.

\section{Design of ODE}
In this section, we first quantify the impact of a local data sample on the performance of global model in terms of convergence rate and inference accuracy. 
Based on the common dominant term in the two analytical expressions, we propose a new data valuation metric for data evaluation and selection in FL (\S\ref{section: Impacts of Local Data Samples}), and develop a practical method to estimate this metric with low extra computation and communication overhead (\S\ref{section: Estimation Method}).
We further design a strategy for the server to coordinate cross-client data selection process, avoiding the potential overlapped data selected and stored by clients (\S\ref{section: Coordination Strategy}). Finally, we summarize  the detailed procedure of {\ttfamily ODE} (\S\ref{section: overall}).

\subsection{Data Valuation Metric}
\label{section: Impacts of Local Data Samples}
We evaluate the impact of a local data sample on FL from the perspectives of convergence rate and inference accuracy, which are two critical aspects for the success of FL. 
The convergence rate quantifies the reduction of loss function in each training round, and determines the communication cost of FL. 
The inference accuracy reflects the effectiveness of a FL model on guaranteeing the quality of service and user experience.
For theoretical analysis, we follow one typical assumption on the FL models, which is widely adopted in the literature~\cite{li2018federated, nguyen2020fast, zhao2018federated}. 
$\vspace{-0.2cm}$
\begin{assumption}(Lipschitz Gradient)
For each client $c\!\in\! C$, the loss function $F_c(w)$ is $L_c$-Lipschitz gradient, \emph{i.e.}, $\parallel\! \nabla_wF_c(w_1)-\nabla_wF_c(w_2)\!\parallel_2\leqslant\! L_c\!\parallel\! w_1-w_2\!\parallel_2$, which implies that the global loss function $F(w)$ is $L$-Lipschitz gradient with $L=\sum_{c\in C}\zeta_cL_c$.
\label{assumption Lipschitz}
$\vspace{-0.2cm}$
\end{assumption}
\noindent Due to the limitation of space, we provide the proofs of all the theorems and lemmas in our technical report \cite{technical_report}.

\textbf{Convergence Rate.}
We provide a lower bound on the reduction of loss function of global model after model aggregation in each communication round. 
\begin{theorem} (Global Loss Reduction)
With Assumption \ref{assumption Lipschitz}, for an arbitrary set of clients $C_t\!\subseteq\! C$ selected by the server in round $t$, the reduction of global loss $F(w)$ is bounded by:
\begin{equation}
    \begin{small}
    \begin{aligned}
       \underbrace{F(w_{\mathrm{fed}}^{t-1})   -F(w_{\mathrm{fed}}^{t})}_{\mathrm{global\ loss\ reduction}}\geqslant&\sum\limits_{c\in C_t}\sum\limits_{i=0}^{m-1}\sum\limits_{(x,y)\in B_c}\big[-\alpha_{c}\underbrace{\parallel\nabla_wl(w_c^{t,i},x,y)\parallel_2^2}_{\mathrm{term}\ 1}\\
       &+\beta_c\underbrace{\langle\nabla_wl(w_c^{t,i},x,y), \nabla_wF(w_{\mathrm{fed}}^{t-1})\rangle}_{\mathrm{term}\ 2}\big],
    \end{aligned}
    \label{equation: global loss reduction}
    \end{small}
\vspace{-0.15cm}
\end{equation}
where $\alpha_c=\frac{L}{2 \zeta_c^t} \cdot \left(\frac{\eta}{|B_c|}\right)^2$ and $\beta_c=\zeta_c^t \cdot\left(\frac{\eta}{|B_c|}\right)$.
\label{theorem 1}
$\vspace{-0.2cm}$
\end{theorem}
Due to the different magnitude orders of coefficients $\alpha_c$ and $\beta_c$\footnote{$\frac{\alpha_c}{\beta_c}\propto\frac{\eta}{|B_c|}\approx10^{-4}$ with common learning rate $10^{-3}$ and storage size $10$.}
and also the values of terms 1 and 2, as is shown in Appendix \ref{appendix: empirical results for claims}, we can focus on the term $2$ (projection of the local gradient of a data sample onto the global gradient) to evaluate a local data sample's impact on the convergence rate.

We next briefly describe how to evaluate data samples based on the term 2. The local model parameter $w_c^{t,i}$ in term $2$ is computed from (\ref{eq: FL model update}), where the gradient $\nabla_w\tilde{F}_c(w_c^{t,i-1})$ depends on the ``cooperation'' of all the stored data samples. Thus, we can formulate the computation of term 2 as a cooperative game~\cite{branzei2008models}, where each data sample represents a player and the utility of the whole dataset is the value of term $2$. 
Within this cooperative game, we can regard the individual contribution of each data sample as its value, and quantify it through leave-one-out~\cite{cook1977detection, koh2017understanding} or Shapley Value \cite{ghorbani2019data, shapley1953quota}. As these methods require multiple model retraining to compute the marginal contribution of each data sample, we propose a one-step look-ahead strategy to approximately evaluate each sample's value by only focusing on the first local training epoch ($m=1$).

\textbf{Inference Accuracy.}
We can assume that the optimal FL model can be obtained by gathering all clients' generated data and conducting CL. Moreover, as the accurate testing dataset and the corresponding testing accuracy are hard to obtain in FL, we use the weight divergence between the models trained through FL and CL to quantify the accuracy of the FL model in each round $t$. With $t\!\rightarrow\!\infty$, we can measure the final accuracy of FL model. 
\begin{theorem}(Model Weight Divergence)
With Assumption \ref{assumption Lipschitz}, for arbitrary set of participating clients $C_t$, we have the following inequality for the weight divergence after the $t^{th}$ training round between the models trained through FL and CL. 
\label{theorem 2}
\begin{equation}
    \begin{small}
    \begin{aligned}
        \parallel w_{\mathrm{fed}}^t-w_{\mathrm{cen}}^{mt}\parallel_2\leqslant&(1+\eta L)^m\parallel w_{\mathrm{fed}}^{t-1}-w_{\mathrm{cen}}^{m(t-1)}\parallel_2\\
        &+\sum\limits_{c\in C_t}\zeta_c^t\left[\eta\sum\limits_{i=0}^{m-1}(1+\eta L)^{m-1-i}G_c(w_c^{t,i})\right],
        \label{equation: model weight divergence}
    \end{aligned}
    \end{small}
\end{equation}
where $G_c(w)=\parallel\nabla_w\tilde{F}_c(w)-\nabla_wF(w)\parallel_2$. 
\end{theorem}
The following lemma further shows the impact of a local data sample on $\parallel\! w_{\mathrm{fed}}^t-w_{\mathrm{cen}}^{mt}\!\parallel_2$ through $G_c(w_c^{t,i})$.
\begin{lemma}
\label{lemma 1}
(Gradient Divergence) For an arbitrary client $c\!\in\! C$, $G_c(w)\!=\parallel\!\nabla \tilde{F}_c(w)-\nabla F(w)\!\parallel_2$ is bounded by:
\begin{equation}
    \begin{footnotesize}
    \begin{aligned}
        G_c(w)\!\leqslant\!\sqrt{\delta+\!\sum\limits_{(x,y)\in B_c}\frac{1}{|B_c|}(\underbrace{\parallel\nabla_wl(w,x,y)\parallel_2^2}_{\mathrm{term\ 1}}-2\underbrace{\langle\nabla_wl(w,x,y), \nabla_wF(w)\rangle}_{\mathrm{term\ 2}})}
        \label{equation: lemma 1},
    \end{aligned}
    \end{footnotesize}
\end{equation}
where $\delta=\parallel\!\nabla_wF(w)\!\parallel_2^2$ is a constant term for all data samples.
$\vspace{-0.2cm}$
\end{lemma}
Intuitively, due to different coefficients, the twofold projection (term $2$) has larger mean and variance than the gradient magnitude (term $1$) among different data samples, which is also verified in our experiments in Appendix \ref{appendix: empirical results for claims}. Thus, we can quantify the impact of a local data sample on $G_c(w)$ and the inference accuracy mainly through term $2$ in (\ref{equation: lemma 1}), which happens to be the same as the term $2$ in the bound of global loss reduction in (\ref{equation: global loss reduction}).

\textbf{Data Valuation}.
\label{section: data valuation}
Based on the above analysis, we define a new data valuation metric in FL, and provide the theoretical understanding as well as intuitive interpretation.

\begin{definition}(Data Valuation Metric) In the $t^{th}$ round, for a client $c\!\in\!C$, the value of a data sample $(x,y)$ is defined as the projection of its local gradient $\nabla l(w,x,y)$ onto the global gradient of the current global model over the unbiased global data distribution:
\begin{equation}
   \begin{small}
    v(x,y)\overset{\mathrm{def}}{=}\langle\nabla_wl(w_{\mathrm{fed}}^t,x,y), \nabla_wF(w_{\mathrm{fed}}^t)\rangle.
    \end{small}
    \label{value definition}
\end{equation}

\end{definition}
Based on this new data valuation metric, once a client receives a new data sample, she can make an online decision on whether to store this sample by comparing the data value of the new sample with those of old samples in storage, which can be easily implemented as a priority queue.

\textit{Theoretical Understanding.}
On the one hand, maximizing the above data valuation metric of the selected data samples is a one-step greedy strategy for minimizing the loss of the updated global model in each training round according to (\ref{equation: global loss reduction}), accelerating model training. 
On the other hand, this metric also improves the inference accuracy of the final global model by narrowing the gap between the models trained through FL and CL, as it reduces the high-weight term of the dominant part in (\ref{equation: model weight divergence}), \emph{i.e.}, $(1+\eta L)^{m-1}G_c(w_k^{t,0})$. 

\textit{Intuitive Interpretation.} 
The proposed data valuation metric guides the clients to select the data samples which not only follow their own local data distribution, but also have similar effect with the global data distribution. 
In this way, the personalized information of local data distribution is preserved and the data heterogeneity across clients is also reduced, which have been demonstrated to improve FL performance~\cite{zawad2021curse,chai2019towards, wang2020tackling,yang2021characterizing}.

\subsection{On-Client Data Selection}
\label{section: Estimation Method}
In practice, it is non-trivial for one client to directly utilize the above data valuation metric for online data selection due to the following two problems: \textit{(1) lack of the latest global model}: due to the partial participation of clients in FL~\cite{jhunjhunwala2022fedvarp}, each client $c\!\in\! C$ does not receive the global FL model $w_{\mathrm{fed}}^{t-1}$ in the rounds that she is not selected, and only has the outdated global FL model from the previous participating round, \textit{i.e.}, $w_{\mathrm{fed}}^{t_{c, \mathrm{last}}-1}$;
\textit{(2) lack of unbiased global gradient}: the accurate global gradient over the unbiased global data distribution can only be obtained by aggregating all the clients' local gradients over their unbiased local data distributions. 
This is hard to achieve because only partial clients participate in each communication round, and the locally stored data distribution could become biased during the on-client data selection process.

We can consider that problem (1) does not affect the online data selection process too much as the value of each data sample remains stable across a few training rounds, which is demonstrated with the experiment results in Appendix \ref{appendix: empirical results for claims}, and thus clients can simply use the old global model for data valuation.

To solve the problem (2), we propose a gradient estimation method.
First, to solve the issue of skew local gradient,
we require each client $c\in C$ to maintain a local gradient estimator $\hat{g}_c$, which will be updated whenever the client receives the $n^{th}$ new data sample $(x,y)$ from the last participating round:
\begin{equation}
    \label{ref:local_gradient}
    \begin{small}
    \hat{g}_c\gets\frac{n-1}{n}\hat{g_c}+\frac{1}{n}\nabla_wl(w_{\mathrm{fed}}^{t_{c,\mathrm{last}}-1},x,y).
    \end{small}
\end{equation}
When the client $c$ is selected to participate in FL at a certain round $t$, the client uploads the current local gradient estimator $\hat{g_c}$ to the server, and resets the local gradient estimator, \emph{i.e.,} $\hat{g_c}\gets 0,n\gets0$, because a new global FL model $w_{\mathrm{fed}}^{t-1}$ is received.
Second, to solve the problem of skew global gradient due to the partial client participation, the server also  maintains a global gradient estimator $\hat{g}^t$, which is an aggregation of the local gradient estimators, $\hat{g}^t=\sum_{c\in C}\zeta_c\hat{g}_c$.
As it would incur high communication cost to collect $\hat{g_c}$ from all the clients, the server only uses $\hat{g}_c$ of the participating clients to update global gradient estimator $\hat{g}^t$ in each round $t$:
\begin{equation}
    \begin{small}
    \hat{g}^t\gets \hat{g}^{t-1}+\sum_{c\in C_t}\zeta_c(\hat{g}_c-\hat{g}_c^{t_{\mathrm{last}}}),
    \label{eq: global gradient}
    \end{small}
\end{equation}
Thus, in each training round $t$, the server needs to distribute both the current global FL model $w^{t-1}_{\mathrm{fed}}$ and the latest global gradient estimator $\hat{g}^{t-1}$ to each selected client $c\!\in\! C_t$, who will conduct local model training, and upload both locally updated model $w_c^{t,m}$ and local gradient estimator $\hat{g_c}$ back to the server.

\textbf{Simplified Version.} 
In both of the local gradient estimation in (\ref{ref:local_gradient}) and data valuation in (\ref{value definition}), for a new data sample, we need to backpropagate the entire model to compute its gradient, which will introduce high computation cost and memory footprint for storing intermediate model outputs. 
To reduce these costs, we only use the gradients of the last few network layers of ML models instead of the whole model, as partial model gradient is also able to reflect the trend of the full gradient, which is also verified in Appendix \ref{appendix: empirical results for claims}. 

\textbf{Privacy Concern.} 
The transmission of local gradient estimators may disclose the local gradient of each client to some extent, which can be avoided by adding Guassian noise to each local gradient estimator before uploading, as in differential privacy~\cite{wei2020federated, dwork2008differential}.


\subsection{Cross-Client Data Storage}
\label{section: Coordination Strategy}
Since the local data distributions of clients may overlap with each other, independently conducting  data selection process for each client may lead to distorted global data distribution.
One potential solution is to divide the global data distribution into several regions, and coordinate each client to store valuable data samples for one specific distribution region, while the union of all  stored data can still follow the unbiased global data distribution.
In this work, we consider the label of data samples as the dividing criterion\footnote{There are some other methods to divide the data distribution, such as K-means \cite{krishna1999genetic} and Hierarchical Clustering~\cite{murtagh2012algorithms}, and our results are independent on these methods.}. 
Thus, before the training process, the server needs to instruct each client the labels and the corresponding quantity of data samples to store. Considering the partial client participation and heterogeneous data distribution among clients, the cross-client coordination strategy need to satisfy the following four desirable properties:

\noindent\textit{(1) Efficient Data Selection:} To improve the efficiency of data selection, the label $y\!\in\! Y$ should be assigned to the clients who generate more data samples with this label, following the intuition that there is a higher probability to select more valuable data samples from a larger pool of candidate data samples.

\noindent\textit{(2) Redundant Label Assignment:} To ensure that all the labels are likely to be covered in each round even with partial client participation, we require each label $y\!\in\!Y$ to be assigned to more than $n^{\mathrm{label}}_y$ clients, which is a hyperparameter decided by the server.

\noindent\textit{(3) Limited Storage:} Due to limited on-device storage, each client $c$ should be assigned to less than $n^{\mathrm{client}}_c$ labels to ensure a sufficient number of valuable data samples stored for each assigned label, and $n^{\mathrm{client}}_c$ is also a hyperparameter decided by the server;

\noindent\textit{(4) Unbiased Global Distribution:} The weighted average of all clients' stored data distribution is expected to be equal to the unbiased global data distribution, \emph{i.e.}, $\tilde{P}(y)=P(y),\forall y\!\in\!Y$.

We formulate the cross-client data storage with the above four properties by representing the coordination strategy as a matrix $D\!\in\!\mathbb{N}^{|C|\times|Y|}$, where $D_{c,y}$ denotes the number of data samples with label $y$ that client $c$ should store. We use matrix $V\!\in\!\mathbb{R}^{|C|\times|Y|}$ to denote the statistical information of each client's generated data, where $V_{c,y}\!=\!v_cP_c(y)$ is the average speed of the data samples with label $y$ generated by client $c$. 
The cross-client coordination strategy can be obtained by solving the following optimization problem, where the condition (1) is formulated as the objective, and conditions (2), (3), and (4) are described by the constraints (\ref{opt: con1}), (\ref{opt: con2}), and (\ref{opt: con3}), respectively:
\begin{subequations}
\begin{align}
    \max_{D}\quad &\sum\limits_{c\in C} \sum\limits_{y\in Y} D_{c,y}\cdot V_{c,y}=\parallel DV\parallel_{1} \label{opt: obj},\\
    \mathrm{s.t.}\quad &\parallel D^{\mathrm{T}}_y\parallel_{0}\geqslant n^{\mathrm{label}}_y, &\forall y\!\in\!Y,\label{opt: con1}\\
    &\parallel D_c\parallel_0\leqslant n_c^{\mathrm{client}}, &\forall c\!\in\!C,\label{opt: con2}\\
    &\parallel D_c\parallel_1=|B_c|,   &\forall c\!\in\!C,\nonumber\\
    &\frac{\sum_{c\in C}D_{c,y}}{\parallel B\parallel_1}=\frac{\sum_{c\in C}V_{c,y}}{\parallel V\parallel_1}, &\forall y\!\in\!Y.\label{opt: con3}
\end{align}
\label{eq: optimization problem}
\end{subequations}
\textbf{Complexity Analysis.}
We can verify that the above optimization problem with $\mathrm{l_0}$ norm is a general convex-cardinality problem, which is NP-hard~\cite{natarajan1995sparse, ge2011note}. To solve this problem, we divide it into two subproblems: (1) decide which elements of matrix $D$ are non-zero, \emph{i.e.}, $S=\{(c,y)|D_{c,y}\ne0\}$, that is to assign labels to clients under the constraints of (\ref{opt: con1}) and (\ref{opt: con2}); (2) determine the specific values of the non-zero elements of matrix $D$ by solving a simplified convex optimization problem: 
\begin{equation}
    \begin{aligned}
      \max_D &\sum_{c \in C, y\in Y, (c,y) \in S}D_{c,y}\cdot V_{c,y}\\
        \mathrm{s.t.} & \sum_{y\in Y, (c, y) \in S}D_{c,y}=|B_c|,   &\forall c\!\in\!C,\\
        &\frac{\sum_{c \in C, (c,y) \in S}D_{c,y}}{\parallel D\parallel_1}=\frac{\sum_{c \in C}V_{c,y}}{\parallel V\parallel_1},  &\forall y\!\in\!Y.
    \end{aligned}
    \label{eq: simplified optimization problem}
\end{equation}
As the number of possible $S$ can be exponential to $|C|$ and $|Y|$, it is still prohibitively expensive to derive the globally optimal solution of $D$ with large $|C|$ (massive clients in FL). 
The classic approach is to replace the non-convex discontinuous $l_0$ norm constraints with the convex continuous $l_1$ norm regularization terms in the objective function~\cite{ge2011note}, which fails to work in our scenario because simultaneously minimizing as many as $|C|+|Y|$ non-differentiable $l_1$ norm in the objective function will lead to high computation and memory costs as well as unstable solutions~\cite{schmidt2007fast, shi2010fast}.
Thus, we propose a greedy strategy to solve this complicated problem.

\textbf{Greedy Cross-Client Coordination Strategy.} 
We achieve the four desirable properties through the following three steps:

\noindent\textit{(1) Information Collection:} 
Each client $c\!\in\!C$ sends the rough information about local data to the server, including the storage capacity $|B_c|$ and data velocity $V_{c,y}$ of each label $y$, which can be obtained from the statistics of previous time periods. Then, the server can construct the vector of storage size $B\!\in\!\mathbb{N}^{|C|}$ and the matrix of data velocity $V\!\in\!\mathbb{R}^{|C|\times|Y|}$ of all clients. 

\noindent\textit{(2) Label Assignment:}
The server sorts labels  according to a non-decreasing order of the total number of clients having this label. 
We prioritize the labels with top rank in label-client assignment, because these labels are more difficult to find enough clients to meet  \textit{Redundant Label Assignment} property.
For each considered label $y\!\in\! Y$ in the rank, there could be multiple clients to be assigned, and the server allocates label $y$ to clients $c$ who generates data samples with label $y$ in a higher data velocity. By doing this, we attempt to satisfy the property of \textit{Efficient Data Selection}.
Once the number of labels assigned to client $c$ is larger than $n_{c}^{\mathrm{client}}$, this client will be removed from the rank due to the \textit{Limited Storage} property.

\noindent\textit{(3) Quantity Assignment:}
With the above two steps, we have decided the non-zero elements of the client-label matrix $D$, \emph{i.e.}, the set $S$. 
To further reduce the computational complexity and avoid the imbalanced on-device data storage for each label, we do not directly solve the simplified optimization problem in (\ref{eq: simplified optimization problem}). Instead, we require each client to divide the storage capacity evenly to the assigned labels, and compute a weight $\gamma_y$ for each label $y\!\in\! Y$ to guarantee that the weighted distribution of the stored data approximates the unbiased global data distribution, \emph{i.e.}, satisfying $\gamma_y\tilde{P}(y)\!=\!P(y)$. Accordingly, we can derive the weight $\gamma_y$ for each label $y$ by setting
\begin{equation}
    \setlength{\abovedisplayskip}{3pt}
    \setlength{\belowdisplayskip}{3pt}
    \begin{small}
    \gamma_y=\frac{P(y)}{\tilde{P}(y)}=\frac{\parallel  V^{\mathrm{T}}_y\parallel_{1} / \parallel  V\parallel_{1}}{\parallel  D^{\mathrm{T}}_y\parallel_{1} / \parallel  D\parallel_{1}}.
    \label{equation: class weight}
    \end{small}
\end{equation}
Thus, each client $c$ only needs to maintain one priority queue with a size $\frac{|B_c|}{\parallel D_c\parallel_{0}}$ for each assigned label. 
In the local model training, each participating client $c\!\in\! C_t$ updates the local model using the weighted stored data samples: 
\begin{equation}
    \setlength{\abovedisplayskip}{3pt}
    \setlength{\belowdisplayskip}{3pt}
    \begin{small}
    w_c^{t,i}\gets w_c^{t,i-1}-\frac{\eta}{\zeta_c}\sum_{(x,y)\in B_c}\gamma_y\nabla_wl(w_c^{t,i-1},x,y),
    \end{small}
    \label{eq: new local model update}
\end{equation}
where $\zeta_c\!=\!\sum_{(x,y)\in B_c}\gamma_y$ denotes the new weight of each client, and the normalized weight of client $c\!\in\!C_t$ for model aggregation in round $t$ becomes $\zeta_c^t=\sum_{c\!\in\!C_t}\frac{\zeta_c}{\sum_{c'\!\in\!C_t}\zeta_{c'}}$.

We illustrate a simple example in Appendix \ref{appendix: simple example} for better understanding of the above procedure.

\textbf{Privacy Concern.}
The potential privacy leakage of uploading rough local information is tolerable in practice, and can be further avoided through Homomorphic Encryption~\cite{acar2018survey}, which enables to sort $k$ encrypted data samples with complexity $O(k\log k^2)$~\cite{hong2021efficient}.

\subsection{Overall Procedure of ODE} 
\label{section: overall}
\begin{figure}
    \centering
    \includegraphics[width=0.45\textwidth]{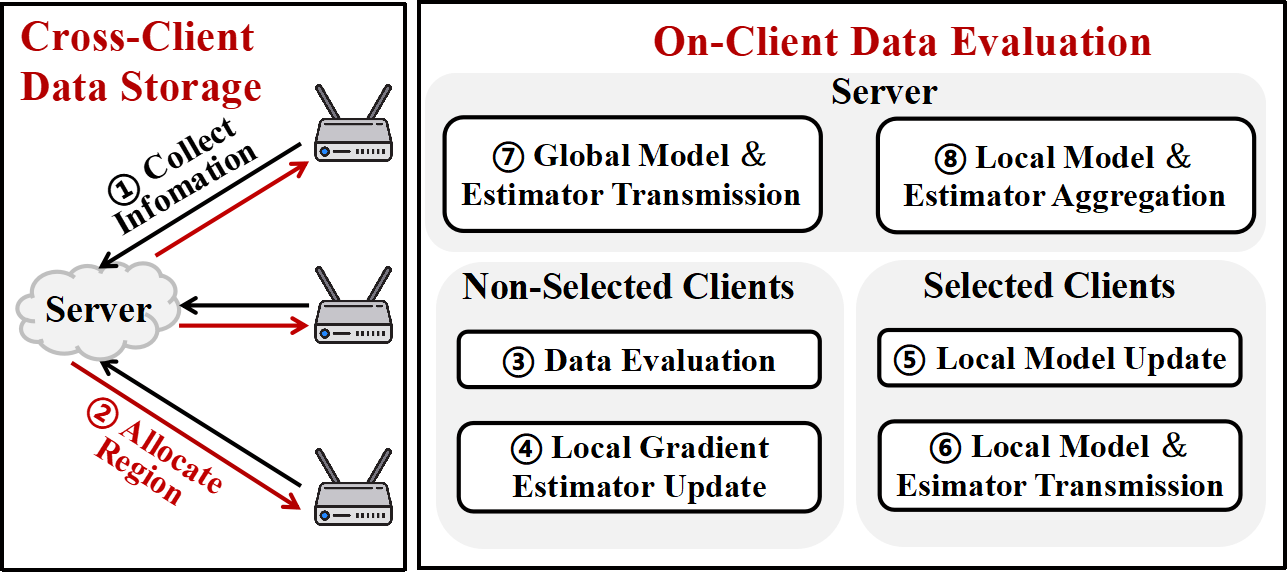}
    \caption{Overview of ODE framework.}
    \Description{Overview of ODE framework.}
    \label{fig: overall design}
    \vspace{-0.6cm}
\end{figure}
{\ttfamily ODE} incorporates cross-client data storage and on-client data evaluation to coordinate mobile devices to store valuable samples, speeding up model training process and improving inference accuracy of the final model. The overall procedure is shown in Figure \ref{fig: overall design}. 

\textbf{Key Idea.} 
Intuitively, the cross-client data storage component coordinates clients to store the high-quality data samples with different labels to avoid highly overlapped data stored by all clients. And the on-client data evaluation component instructs each client to select the data having similar gradient with the global data distribution, which reduces the data heterogeneity among clients while also preserves personalized information. 

\textbf{Cross-Client Data Storage.}
Before the FL process, the central server collects data distribution information and storage capacity from all clients (\ding{172}), and solving the optimization problem in (9) through our greedy approach (\ding{173}).

\textbf{On-Client Data Evaluation.}
During the FL process, clients train the local model in participating rounds, and utilize idle computation and memory resources to conduct on-device data selection in non-participating rounds. 
In the $t^{th}$ training round, non-selected clients, selected clients and the server perform different operations:

\noindent$\bullet\ $\textbf{Non-selected clients}: 
\textit{Data Evaluation} (\ding{174}): each non-selected client $c\!\in\! C\setminus C_t$ continuously evaluates and selects the data samples according to the data valuation metric in (\ref{value definition}), within which the clients use the estimated global gradient received in last participation round instead of the accurate global one. 
\textit{Local Gradient Estimator Update} (\ding{175}): the client also continuously updates the local gradient estimator $\hat{g}_c$ using (\ref{ref:local_gradient}). 

\noindent$\bullet\ $\textbf{Selected Clients}: 
\textit{Local Model Update} (\ding{176}): after receiving the new global model $w_\mathrm{fed}^{t-1}$ and new global gradient estimator $\hat{g}^{t-1}$ from the server, each selected client $c\!\in\! C_t$ performs local model updates using the stored data samples by (\ref{eq: new local model update}). 
\textit{Local Model and Estimator Transmission} (\ding{177}): each selected client sends the updated model $w_c^{t,m}$ and local gradient estimator $\hat{g}_c$ to the server. The local estimator $\hat{g}_c$ will be reset to $0$ for approximating local gradient of the newly received global model $w_\mathrm{fed}^{t-1}$.

\noindent$\bullet\ $ \textbf{Server}:
\textit{Global Model and Estimator Transmission (\ding{178})}: At the beginning of each training round, the server distributes the global model $w_{\mathrm{fed}}^{t-1}$ and the global gradient estimator $\hat{g}^{t-1}$ to the selected clients. 
\textit{Local Model and Estimator Aggregation} (\ding{179}): at the end of each training round, the server collects the updated local models $w_c^{t,m}$ and local gradient estimators $\hat{g}_c$ from participating clients $c\!\in\! C_t$, which will be aggregated to obtain a new global model by (\ref{eq: FL model aggregation}) and a new global gradient estimator by (\ref{eq: global gradient}).


\section{Evaluation}
\label{section: evaluation}
In this section, we first introduce experiment setting, baselines and evaluation metrics.
Second, we present the overall performance of {\ttfamily ODE} and baselines on model training speedup and inference accuracy improvement, as well as the memory footprint and evaluation delay. Next, we show the robustness of {\ttfamily ODE} against various environment factors. 
We also show \textit{the individual and integrated impacts of limited storage and streaming data on FL} to show our motivation, and analyze the individual \textit{effect of different components} of {\ttfamily ODE}, which are shown in Appendix \ref{appendix: motivating experiments} and \ref{appendix: component-wise analysis} due to the limited space.

\subsection{Experiment Setting}
\label{section: experiment setting}
\textbf{Tasks, Datasets and ML Models.} 
To demonstrate the {\ttfamily ODE}'s good performance and generalization across various tasks, datasets and ML models. we evaluate {\ttfamily ODE} on one \textit{synthetic} dataset, two \textit{real-world} datasets and one \textit{industrial} dataset, all of which vary in data quantities, distributions and model outputs, and cover the tasks of Synthetic Task (ST), Image Classification (IC), Human Activity Recognition (HAR) and mobile Traffic Classification (TC). The statistics of the tasks are summarized in Table \ref{tab: statistical information of tasks and datasets}, and introduced in details in Appendix \ref{appendix: introduction to tasks and datasets}.
\begin{table*}[]
    \vspace{-0.2cm}  
    \setlength{\abovecaptionskip}{-0.1cm}   
    \setlength{\belowcaptionskip}{0cm}   
    \centering
    \begin{tabular}{c|ccccc|ccccc}
    \toprule[1.5pt]
    \textbf{Tasks} & \textbf{Datasets} & \textbf{Models} & \textbf{$\#$Samples} & \textbf{$\#$Labels}& \textbf{$\#$Devices} & \bm{$n_y^{\mathrm{label}}$}  & \bm{$\frac{|C_t|}{|C|}$} & \bm{$\eta$} & \bm{$m$} & \bm{$|B_c|$}\\
    \hline
    ST & Synthetic Dataset~\cite{caldas2018leaf} & LogReg & $1,016,442$ 
    & $10$ & $200$ & $5$ & $5\%$ & $1e^{-4}$ & $5$ & $10$\\
    IC & Fashion-MNIST~\cite{xiao2017/online} & LeNet~\cite{lecun1989handwritten}& $70,000$ & $10$ & $50$ & $5$ & $10\%$ & $1e^{-3}$ & $5$ & $5$ \\
    HAR & HARBOX~\cite{ouyang2021clusterfl} & Customized DNN & $34,115$ & $5$ & $120$ & $5$ & $10\%$ & $1e^{-3}$ & $5$ & $5$\\
    TC & Industrial Dataset & Customized CNN & $37,853$ & $20$ & $30$  & $5$ & $20\%$ & $5e^{-3}$ & $5$ & $10$\\
    \bottomrule[1.2pt]
    \end{tabular}
    \caption{Information of different tasks, datasets, models and default experiment settings\protect\footnotemark[6].}
    \label{tab: statistical information of tasks and datasets}
\end{table*}

\textbf{Parameters Configurations.}
The main configurations are shown in Table \ref{tab: statistical information of tasks and datasets} and other configurations like the training optimizer and velocity of on-device data stream are presented in Appendix \ref{appendix: configurations}.

\textbf{Baselines.} In our experiments, we compare two versions of {\ttfamily ODE}, \textit{ODE-Exact} (using exact global gradient) and \textit{ODE-Est} (using estimated global gradient), with four categories of data selection methods, including random sampling methods (\textit{RS}), importance-sampling based methods for CL (\textit{HL} and \textit{GN}), previous data selection methods for FL (\textit{FB} and \textit{SLD}) and the ideal case with unlimited on-device storage (\textit{FD}). These methods are introduced in details in Appendix \ref{appendix: introduction to baselines}.


\textbf{Metrics for Training Performance.} We use two metrics to evaluate the performance of each method: 
(1) \textit{Time-to-Accuracy Ratio}: 
we measure the training speedup of global model by the ratio of training time of \textit{RS} 
and the considered method to reach the same target accuracy, which is set to be the final inference accuracy of \textit{RS}. As the time of one communication round is usually fixed in practical FL scenario, we can quantify the training time with the number of communicating rounds.
(2) \textit{Final Inference Accuracy}: we evaluate the inference accuracy of the final global model on each device's testing data and report the average accuracy for evaluation.

\subsection{Overall Performance}
\begin{figure*}
    \begin{minipage}[t]{0.65\textwidth}
        \centering
        \vspace{-0.1cm}
        \begin{tabular}{p{0.6cm}<{\centering}|>{\columncolor{red!15}}p{0.8cm}<{\centering}|p{0.8cm}<{\centering}p{0.8cm}<{\centering}|p{0.8cm}<{\centering}p{0.8cm}<{\centering}|>{\columncolor{blue!25}}c >{\columncolor{blue!15}}c|>{\columncolor{orange!15}}c}
        \hline
        \multirow{2}{*}{\textbf{Task}} & \multicolumn{8}{c}{\textbf{Model Training Speedup}}\\
        \cline{2-9}
        & \textbf{RS} &
        \textbf{HL} &
        \textbf{GN} & 
        \textbf{FB} & \textbf{SLD} & 
        \textbf{ODE-Exact} & 
        \textbf{ODE-Est} & 
        \textbf{FD} \\
        \hline
         ST & $1.0\times$ & $-$ & $4.87\times$ & $-$ & $4.08\times$ & $9.52\times$ & $5.88\times$ & $2.67\times$\\
        IC & $1.0\times$ & $-$ & $-$ & $-$ & $-$ & $1.35\times$ & $1.20\times$ & $1.01\times$\\
        HAR & $1.0\times$ & $-$ & $-$ & $-$ & $-$ & $2.22\times$ & $1.55\times$ & $4.76\times$\\
        TC & $1.0\times$ & $-$ & $-$ & $-$ & $-$ & $2.51\times$ & $2.50\times$ & $3.92\times$\\
        \hline
        \textbf{Task} & \multicolumn{8}{c}{\textbf{Inference Accuracy}}\\
        \hline
        ST & $79.56\%$ & $78.44\%$ & $83.28\%$ & $78.56\%$ & $82.38\%$ & $87.12\%$ & $82.80\%$ & $88.14\%$\\
        IC & $71.31\%$ & $51.95\%$ & $41.45\%$ & $60.43\%$ & $69.15\%$
        & $72.71\%$ & $72.70\%$ & $71.37\%$\\
        HAR & $67.25\%$ & $48.16\%$ & $51.02\%$ & $48.33\%$ & $56.24\%$ & $73.63\%$ & $70.39\%$ & $77.54\%$\\
        TC & $89.3\%$ & $69.00\%$ & $69.3\%$ & $72.19\%$ & $72.30\%$ & $95.3\%$ & $95.30\%$ & $96.00\%$\\
        \hline
        \end{tabular}
        \vspace{0cm}
        \captionof{table}{{\ttfamily ODE}'s improvements on  model training speedup and inference accuracy. 
        The symbol '$-$' means that the method fails to reach the target accuracy.}
        \label{tab: overall performance of ODE on public datasets}
        \vspace{-0.6cm}
    \end{minipage}
    \begin{minipage}[t]{0.33\textwidth}
            \vspace{-0.2cm}  
            \centering \includegraphics[width=0.45\textwidth]{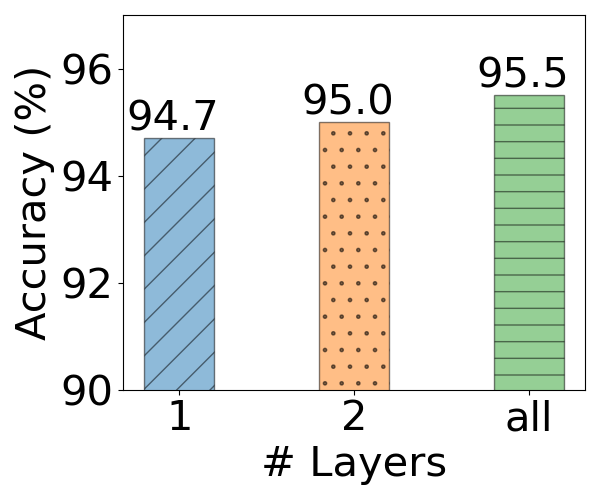} \includegraphics[width=0.45\textwidth]{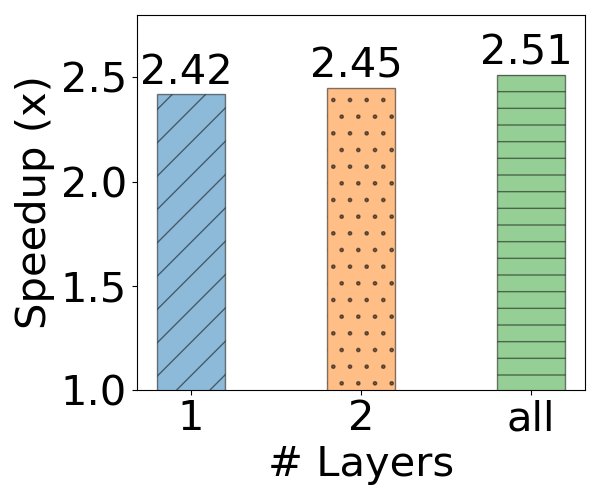} \includegraphics[width=0.45\textwidth]{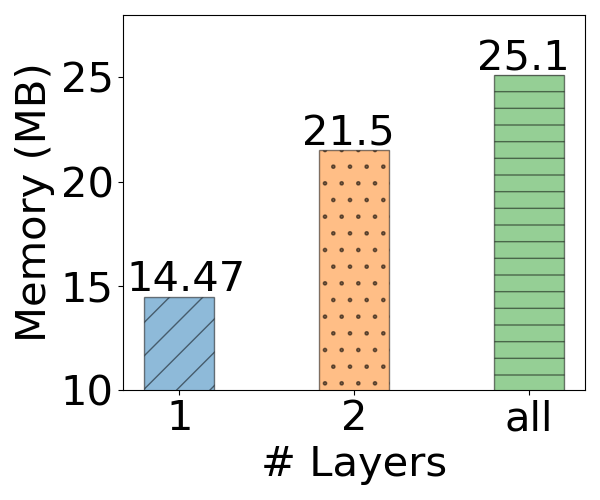} \includegraphics[width=0.45\textwidth]{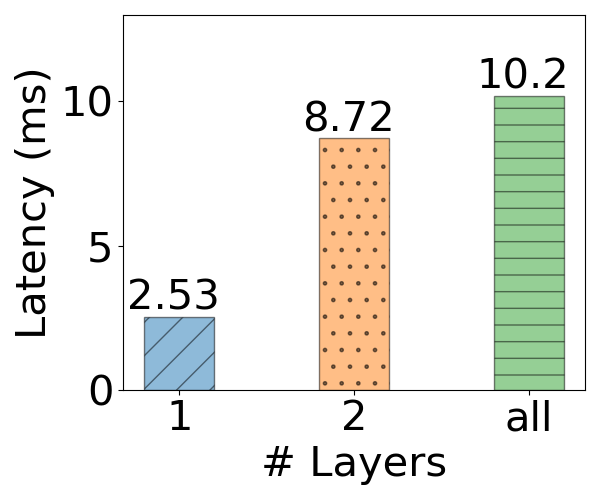}
            
            \vspace{-0.5cm}  
            \caption{Performance and cost of simplified {\ttfamily ODE} with different $\#$model layers.}
            \Description{Performance and cost of simplified {\ttfamily ODE} with different $\#$model layers.}
            \label{fig: simplified estimation}
            \vspace{-0.86cm}  
    \end{minipage}
\end{figure*}

We compare the performance of {\ttfamily ODE} with four baselines on all the datasets, and show the results in Table \ref{tab: overall performance of ODE on public datasets}.

\textit{{\ttfamily ODE} significantly speeds up the model training process.}
We observed that {\ttfamily ODE} improves time-to-accuracy performance over the existing data selection methods on all the four datasets. Compared with baselines, {\ttfamily ODE} achieves the target accuracy $5.88\times\!\!\sim\!\!9.52\times$ faster on ST; $1.20\times\!\!\sim\!\!1.35\times$ faster on IC;  $1.55\times\!\!\sim\!\!2.22\times$ faster on HAR; $2.5\times$ faster on TC. 
Also, we observe that the largest speedup is achieved on the datast ST, because the high non-i.i.d degree across clients and large data divergence within clients leave a great potential for {\ttfamily ODE} to reduce data heterogeneity and improve training process through data selection.

\textit{{\ttfamily ODE} largely improves the final inference accuracy.} 
Table \ref{tab: overall performance of ODE on public datasets} shows that in comparison with baselines with the same storage, {\ttfamily ODE} enhances final accuracy on all the datasets, achieving $3.24\%\!\!\sim\!\!7.56\%$ higher on ST, $3.13\%\!\!\sim\!\!6.38\%$ increase on HAR, and around $6\%$ rise on TC. We also notice that {\ttfamily ODE} has a marginal accuracy improvement ($\approx\!1.4\%$) on IC, because the FashionMNIST has less data variance within each label, and a randomly selected subset is sufficient to represent the entire data distribution for model training.

\textit{Importance-based data selection methods perform poorly.}
Table \ref{tab: overall performance of ODE on public datasets} shows that these methods even cannot reach the target final accuracy on tasks IC, HAR and TC, as these datasets are collected from real world and contain noise data, making such importance sampling methods fail to work~\cite{DBLP:conf/mobisys/ShinLLL22, li2021sample}. 

\textit{Previous data selection methods for FL outperform importance based methods but worse than {\ttfamily ODE}}. 
As is shown in Table \ref{tab: overall performance of ODE on public datasets}, \textit{FedBalancer} and \textit{SLD} perform better than \textit{HL} and \textit{GN}, but worse than \textit{RS} in a large degree, which is different from the phenomenon in traditional settings~\cite{katharopoulos2018not, li2021sample, DBLP:conf/mobisys/ShinLLL22}. 
This is because (1) their noise reduction steps, such as removing samples with top loss or gradient norm, highly rely on the complete statistical information of full dataset, and (2) their on-client data valuation metrics fail to work for the global model training in FL, as discussed in \S\ref{section: introduction}.

\textit{Simplified {\ttfamily ODE} reduces computation and memory costs significantly with little performance degradation.}
We conduct another two experiments which consider only the last 1 and 2 layers (5 layers in total) for data valuation on the industrial TC dataset. Empirical results shown in Figure \ref{fig: simplified estimation} demonstrate that the simplified version reduces as high as $44\%$ memory and $83\%$ time delay, with only $1\%$ and $0.1\times$ degradation on accuracy and speedup. 

\textit{{\ttfamily ODE} introduces small extra memory footprint and data processing delay during data selection process}. 
\begin{table}
    \centering
    \begin{tabular}{c|ccccc}
    \toprule[1.5pt]
    \multirow{2}{*}{\textbf{Task}} & \multicolumn{5}{c}{\textbf{Memory Footprint (MB)}}\\
    & \textbf{RS} & \textbf{HL} & \textbf{GN} & \textbf{ODE-Est} & \textbf{ODE-Simplified}\\
    \hline
    IC & 1.70 & 11.91 & 16.89 & 18.27 & 16.92\\
    HAR & 1.92 & 7.27 & 12.23 & 13.46 & 12.38\\
    TC & 0.75 & 10.58 & 19.65  & 25.15 & 14.47\\
    \midrule[1pt]
    \textbf{Task} & \multicolumn{5}{c}{\textbf{Evaluation Time (ms)}}\\
    \hline
    IC & 0.05 & 11.1 & 21.1 & 22.8 & 11.4\\
    HAR & 0.05 & 0.36 & 1.04 & 1.93 & 0.53\\
    TC & 0.05 & 1.03 & 9.06 & 9.69 & 1.23\\
    \bottomrule[1.2pt]
    \end{tabular}
    \caption{The memory footprint and evaluation delay per sample valuation of baselines on three real-world tasks.}
    \label{tab: memory footprint and evaluation time}
    \vspace{-0.6cm} 
\end{table}
Empirical results in Table \ref{tab: memory footprint and evaluation time} demonstrate that simplified {\ttfamily ODE} brings only tiny evaluation delay and memory burden to mobile devices ($1.23$ms and $14.47$MB for TC task), and thus can be applied to practical network scenario.

\subsection{Robustness of ODE}
\label{section: Effectiveness and Robustness of ODE}

In this subsection, we mainly compare the robustness of {\ttfamily ODE} and previous methods to various factors in industrial environments, such as the number of local training epoch $m$, client participation rate $\frac{|C_t|}{|C|}$, storage capacity $|B_c|$, mini-batch size and data heterogeneity across clients, on the industrial TC dataset. 

\begin{figure}
    \vspace{-0.4cm}
    \centering
    \subfigure[Epoch=2]{
        \includegraphics[height=2.6cm]{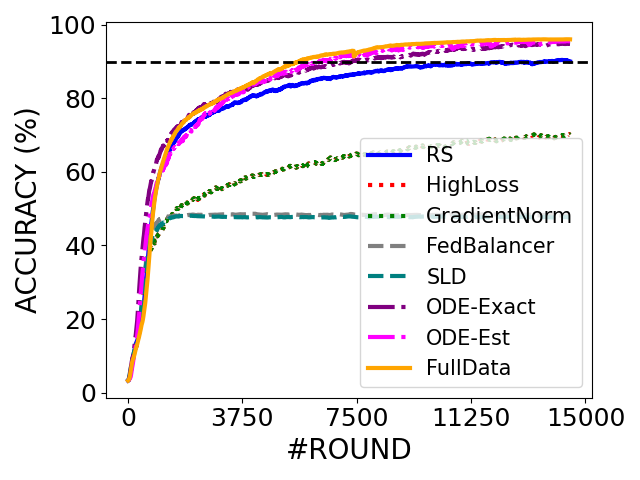}
        \label{fig: epoch=2}
        }
    \subfigure[Epoch=10]{
        \includegraphics[height=2.6cm]{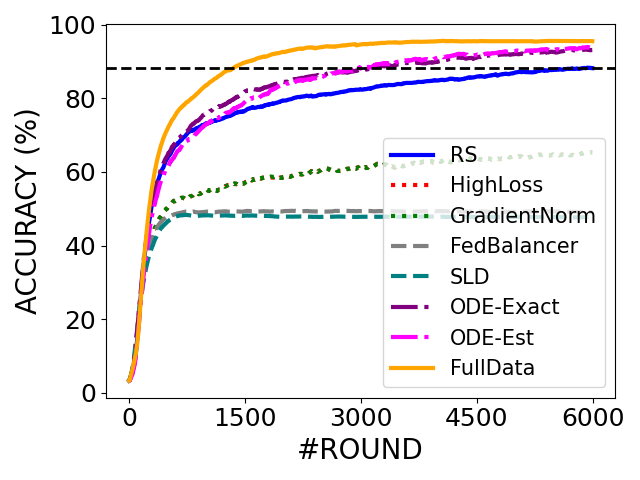}
        \label{fig: epoch=10}
        }
    \vspace{-0.5cm}  
    \caption{The training process of different sampling methods with various numbers of local epoch.}
    \Description{The training process of different sampling methods with various numbers of local epoch.}
    \label{fig: epoch}
    \vspace{-0.3cm}
\end{figure}

\textbf{Number of Local Training Epoch.} 
Empirical results shown in Figure \ref{fig: epoch} demonstrate that 
{\ttfamily ODE} can work with various local training epoch numbers $m$. With $m$ increasing, both of \textit{ODE-Exact} and \textit{ODE-Est} achieve higher final inference accuracy than existing methods with same setting.



\begin{figure}
    \vspace{-0.2cm}  
    \centering
    \subfigure[Training Speedup]{
        \includegraphics[height=2.3cm]{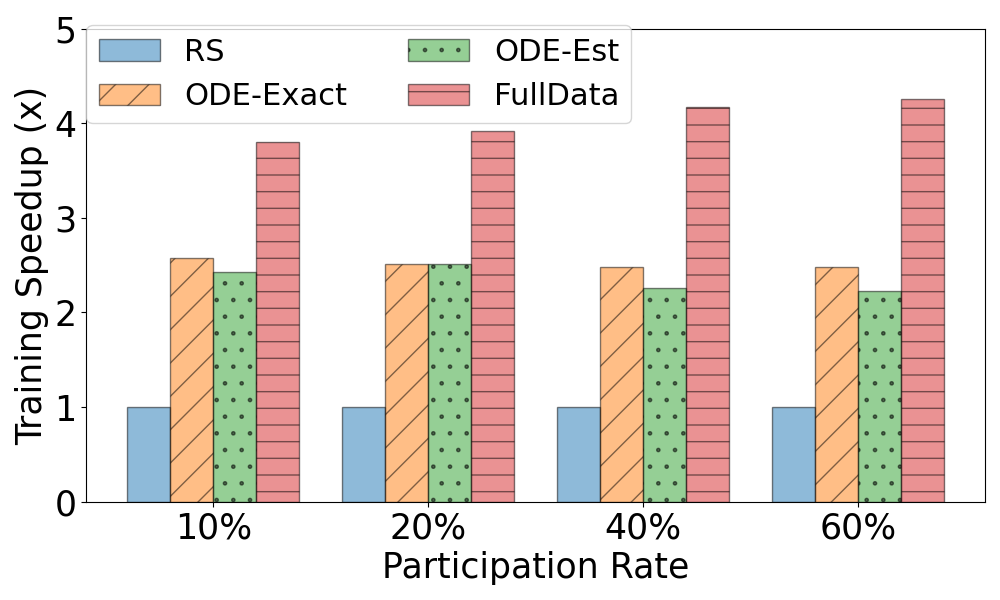}
    }
    \subfigure[Final Accuracy]{
        \includegraphics[height=2.3cm]{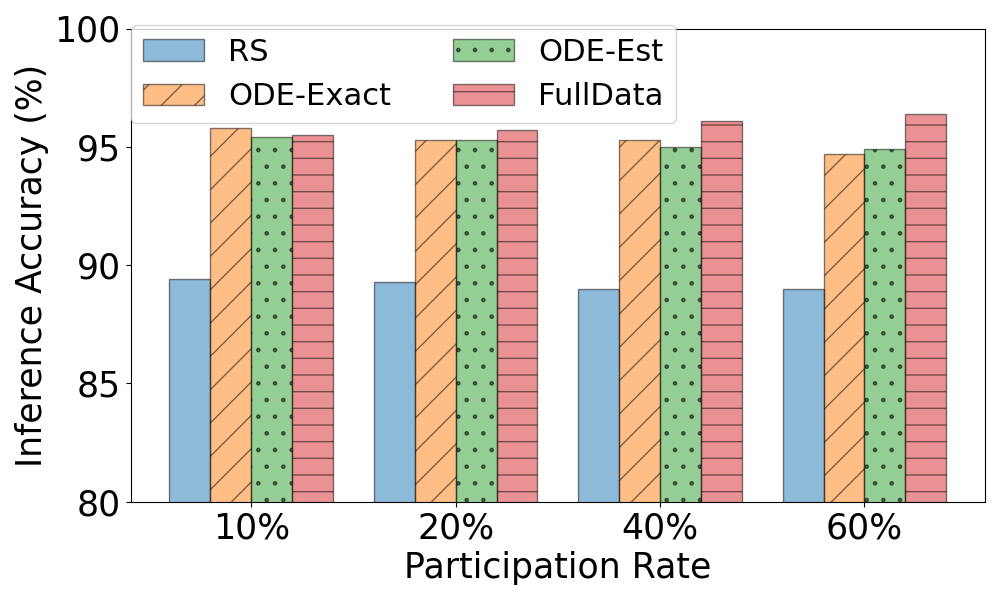}
    }
    \vspace{-0.5cm} 
    \caption{The performance of different data selection methods with various participation rates. 
    }
    \Description{The performance of different data selection methods with various participation rates.}
    \label{fig: participation rate}
    \vspace{-0.4cm}
\end{figure}

\textbf{Participation Rate.} 
The results in Figure \ref{fig: participation rate} demonstrate that {\ttfamily ODE} can improve the FL process significantly even with small participation rate, accelerating the model training $2.57\times$ and increasing the inference accuracy by $6.6\%$.
This demonstrates the practicality of {\ttfamily ODE} in the industrial environment, where only a small proportion of mobile devices could be ready to participate in each FL round.

\textbf{Other Factors.} We also demonstrate the remarkable robustness of {\ttfamily ODE} to \textbf{device storage capacity}, \textbf{mini-batch size} and \textbf{data heterogeneity across clients} compared with previous methods, which are fully presented in the technical report~\cite{technical_report}.

\section{Related Works}
\label{section: related work}
\textbf{Federated Learning} is a distributed learning framework that aims to collaboratively learn a global model over the networked devices' data under the constraint that the data is stored and processed locally
\cite{li2020federated,mcmahan2017communication}. 
Existing works mostly focus on how to overcome the data heterogeneity problem \cite{chai2019towards, zawad2021curse, DBLP:conf/ijcai/ChoMJSD22, wolfrath2022haccs}, reduce the communication cost \cite{khodadadian2022federated, khodadadian2022federated, hou2021fedchain, yang2020federated}, select important clients~\cite{cho2022towards, nishio2019client, li2021hermes, li2021federated} or train a personalized model for each client \cite{wu2021hierarchical, fallah2020personalized}. Despite that a few works consider the problem of online and continuous FL \cite{wang2022federated, chen2020asynchronous, yoon2021federated}, they did not consider the device properties of \textit{limited on-device storage} and \textit{streaming networked data}.

\textbf{Data Selection.} 
In FL, selecting data from streaming data can be seen as sampling batches of data from its distribution, which is similar to mini-batch SGD. To improve the training process of SGD, existing methods quantify the importance of each data sample (such as loss \cite{shrivastava2016training, schaul2015prioritized}, gradient norm \cite{johnson2018training, zhao2015stochastic}, uncertainty \cite{chang2017active, wu2017sampling}, data shapley~\cite{ghorbani2019data} and representativeness~\cite{mirzasoleiman2020coresets, wang2021fair}) and leverage importance sampling or priority queue to select training samples. 
The previous literature~\cite{DBLP:conf/mobisys/ShinLLL22, li2021sample} on data selection in FL simplify conducts the above data selection methods on each client individually for local model training without considering the global model. And all of them require either access to all the data or multiple inspections over the data stream, which are not satisfied in the mobile network scenarios.
$\vspace{-0.2cm}$
\section{Conclusion}
In this work, we identify two key properties of networked FL: \textit{limited on-device storage} and \textit{streaming networked data}, which have not been fully explored in the literature. 
Then, we present the design, implementation and evaluation of {\ttfamily ODE}, which is an online data selection framework for FL with limited on-device storage, consisting of two components: on-device data evaluation and cross-device collaborative data storage. 
Our analysis show that {\ttfamily ODE} improves both convergence rate and inference accuracy of the global model, simultaneously.
Empirical results on three public and one industrial datasets demonstrate that {\ttfamily ODE} significantly outperforms the state-of-the-art data selection methods in terms of training time, final accuracy and robustness to various factors in industrial environments. 

\clearpage
\begin{acks}
This work was supported in part by National Key R$\&$D Program of China No. 2020YFB1707900, in part by China NSF grant No. 62132018, U2268204, 62272307 61902248, 61972254, 61972252, 620252\\
04, 62072303, in part by Shanghai Science and Technology fund 20PJ1407900, in part by Huawei Noah's Ark Lab NetMIND Research Team, in part by Alibaba Group through Alibaba Innovative Research Program, and in part by Tencent Rhino Bird Key Research Project. The opinions, findings, conclusions, and recommendations expressed in this paper are those of the authors and do not necessarily reflect the views of the funding agencies or the government. Zhenzhe Zheng is the corresponding author.
\end{acks}

\bibliographystyle{ACM-Reference-Format}
\bibliography{main.bib}

\clearpage

\appendix
\section{Simple Example}
\label{appendix: simple example}
\begin{figure}[H]
    \centering
    \vspace{-0.1cm}  
    \includegraphics[width=0.4\textwidth]{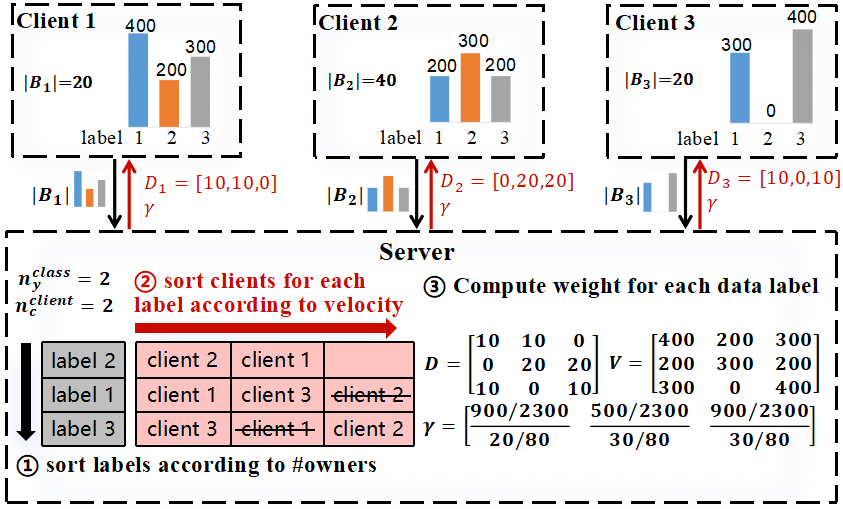}
    \vspace{-0.3cm}
    \caption{A simple example to illustrate the greedy coordination. 
    \ding{172} Sort labels according to $\#$owners and obtain label order $\{2,1,3\}$; \ding{173} Sort and allocate clients for each label under the constraint of $n_c^\mathrm{client}$ and $n_y^\mathrm{class}$;
    \ding{174} Obtain coordination matrix $D$ and compute class weight $\gamma$ according to (\ref{equation: class weight}).
    }
    \Description{A simple example to illustrate the greedy coordination. 
    \ding{172} Sort labels according to $\#$owners and obtain label order $\{2,1,3\}$; \ding{173} Sort and allocate clients for each label under the constraint of $n_c^\mathrm{client}$ and $n_y^\mathrm{class}$;
    \ding{174} Obtain coordination matrix $D$ and compute class weight $\gamma$ according to (\ref{equation: class weight}).
    }
    \label{fig: simple example}
    \vspace{-0.5cm}
\end{figure}

In Figure \ref{fig: simple example}, We use a simple example to illustrate our proposed greedy solution for the cross-device collaborative data selection described in \S\ref{section: Coordination Strategy}.

\section{Empirical results for claims.}
\label{appendix: empirical results for claims}
\begin{figure}[H]
    \centering
    \subfigure[Term 1 vs Term 2]{
    \includegraphics[width=0.14\textwidth]{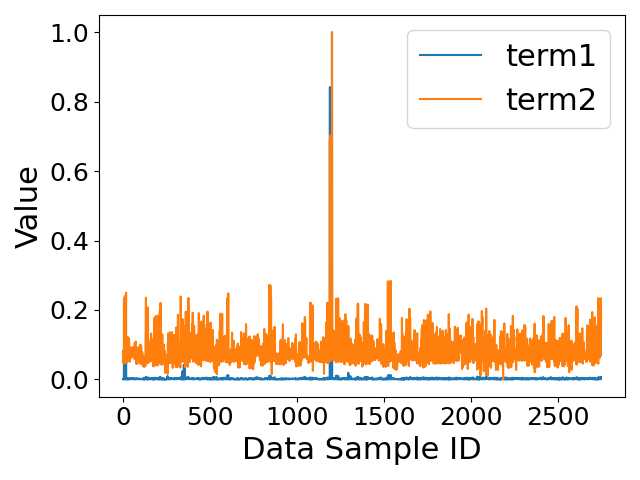}
    \label{fig: assumption term}
    }
    \subfigure[Round 0 vs Round 30]{
    \includegraphics[width=0.14\textwidth]{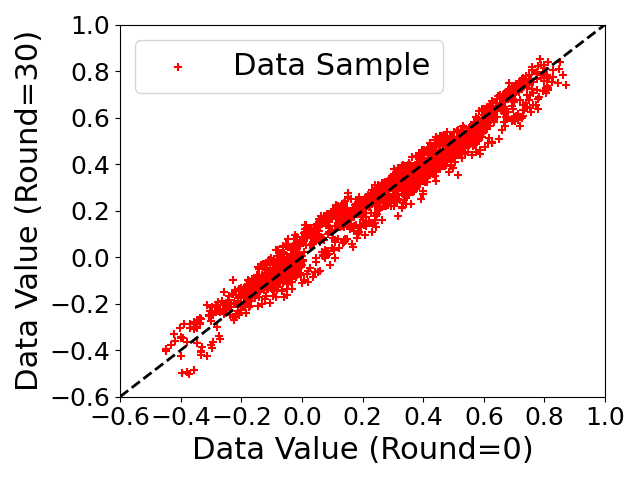}
    \label{fig: assumption value}
    }
    \subfigure[Full model vs Last layer]{
    \includegraphics[width=0.14\textwidth]{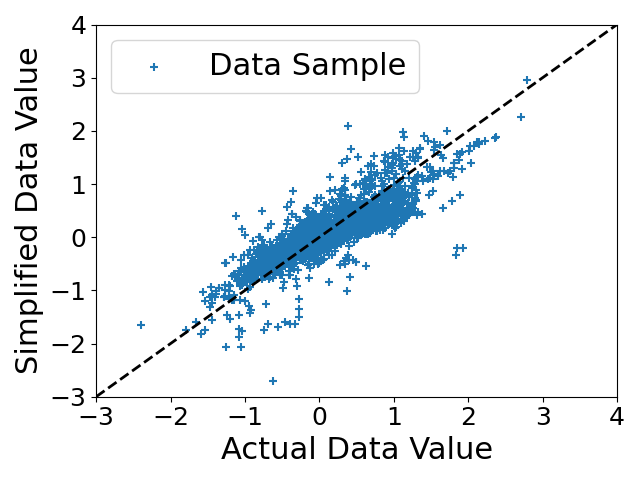}
    \label{fig: assumption simplified}
    }
    \caption{Empirical results to support some claims, and the experiment setting could be found in \S\ref{section: experiment setting}. 
    (a): comparison of the normalized values of two terms in (\ref{equation: global loss reduction}) and (\ref{equation: lemma 1}). 
    (b): comparison of the data values computed in round 0 and round 30. 
    (c): comparison of the data values computed using gradients of full model layers and the last layer.}
    \Description{Empirical results to support some claims, and the experiment setting could be found in \S\ref{section: experiment setting}. 
    (a): comparison of the normalized values of two terms in (\ref{equation: global loss reduction}) and (\ref{equation: lemma 1}). 
    (b): comparison of the data values computed in round 0 and round 30. 
    (c): comparison of the data values computed using gradients of full model layers and the last layer.}
    \label{fig: assumption}
\end{figure}
Empirical results to support some  claims mentioned before, and the experiment setting could be found in \S\ref{section: experiment setting}. 
Figure \ref{fig: assumption term}: comparison of the normalized values of two terms in (\ref{equation: global loss reduction}) and (\ref{equation: lemma 1}). 
Figure \ref{fig: assumption value}: comparison of the data values computed in round 0 and round 30. 
Figure \ref{fig: assumption simplified}: comparison of the data values computed using gradients of full model layers and the last layer.

\section{Experiments}

\subsection{Tasks and Datasets}
\label{appendix: introduction to tasks and datasets}
\textbf{Synthetic Task.} The synthetic dataset we used is proposed in LEAF benchmark~\cite{caldas2018leaf} and is also described in details in~\cite{li2018federated}. It contains $200$ clients and 1 million data samples, and a Logistic Regression model is trained for this 10-class task.
    
\textbf{Image Classification.} Fashion-MNIST~\cite{xiao2017/online} contains $60,000$ training images and $10,000$ testing images, which are divided into 50 clients according to labels~\cite{mcmahan2017communication}. We train LeNet~\cite{lecun1989handwritten} for the 10-class image classification.
    
\textbf{Human Activity Recognition.} HARBOX~\cite{ouyang2021clusterfl} is the 9-axis OMU dataset collected from 121 users' smartphones in a crowdsourcing manner, including 34,115 data samples with $900$ dimension. Considering the simplicity of the dataset and task, a lightweight customized DNN with two dense layers followed by a SoftMax layer is deployed for this 5-class human activity recognition task~\cite{lipyramidfl}.
    
\begin{figure}
    \vspace{-0.3cm}  
    \centering
    \begin{minipage}{0.2\textwidth}
        \centering
        \includegraphics[height=2.3cm]{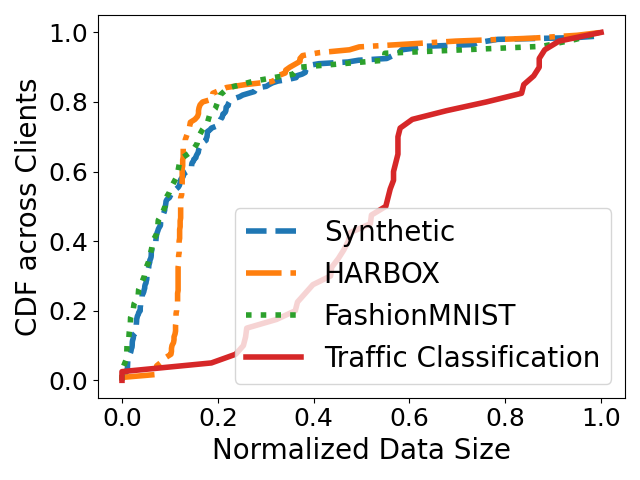}
        \vspace{-0.3cm}
        \caption{Unbalanced data quantity of clients.}
        \Description{Unbalanced data quantity of clients.}
        \label{fig: distribution of public datasets}
    \end{minipage}
    \begin{minipage}{0.27\textwidth}
        \centering
        \includegraphics[height=2.3cm]{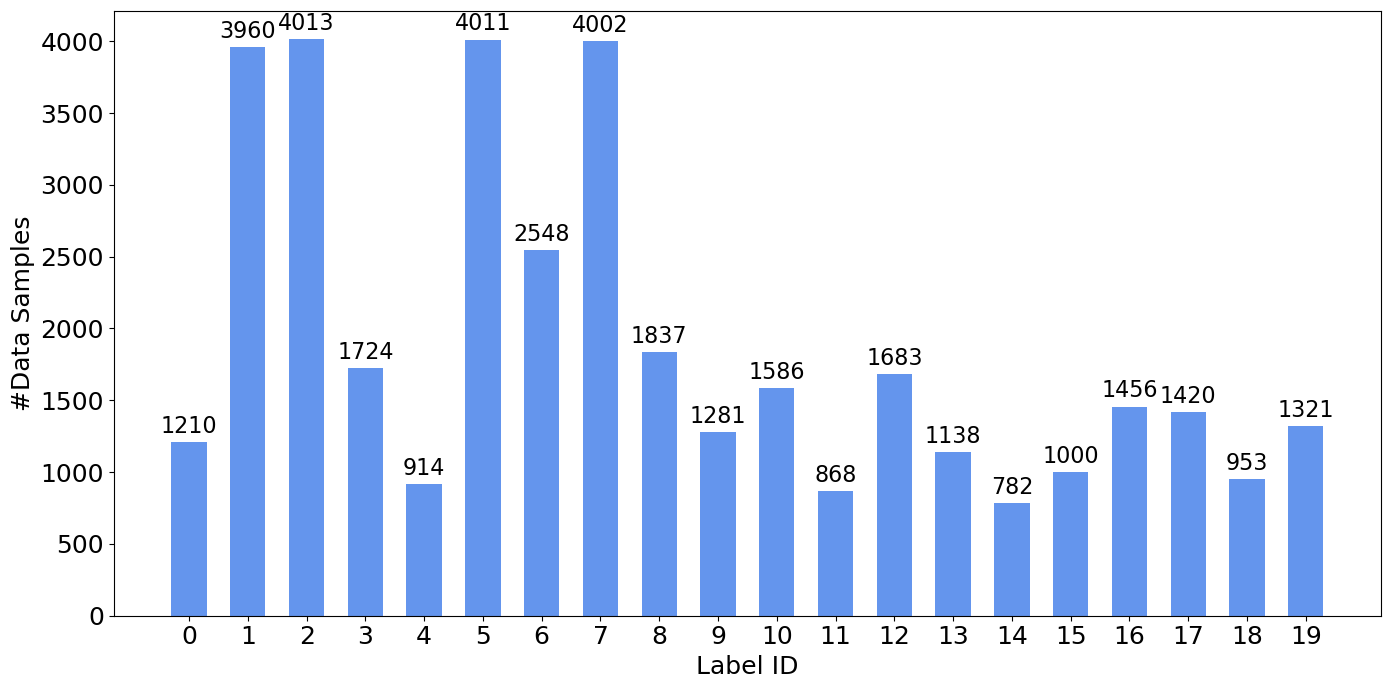}
        \vspace{-0.3cm}
        \caption{Data distribution in traffic classification dataset.}
        \Description{Data distribution in traffic classification dataset.}
        \label{fig: label distribution}
    \end{minipage}
    \vspace{-0.4cm}
\end{figure}

\textbf{Traffic Classification.} 
The industrial dataset about the task of mobile application classification is collected by our deployment of $30$ ONTs (optimal network terminal) in a simulated network environment from May 2019 to June 2019. Generally, the dataset contains more than $560,000$ data samples and has more than $250$ applications as labels, which cover the application categories of videos (such as YouTube and TikTok), games (such as LOL and WOW), files downloading (such as AppStore and Thunder) and communication (such as WhatsApp and WeChat). We manually label the application of each data sample. The model we applied is a CNN consisting of 4 convolutional layers with kernel size $1\times 3$ to extract features and 2 fully-connected layers for classification, which is able to achieve $95\%$ accuracy through CL and satisfy the on-device resource requirement due to the small number of model parameters. To reduce the training time caused by the large scale of dataset, we randomly select $20$ out of $250$ applications as labels with various numbers of data samples, whose distribution is shown in Figure \ref{fig: label distribution}.

\subsection{Configurations}
\label{appendix: configurations}
For all the experiments, we use SGD as the optimizer and decay the learning rate per $100$ rounds by $\eta_{\mathrm{new}}=0.95\times\eta_{\mathrm{old}}$. To simulate the setting of streaming data, we set the on-device data velocity to be $v_c=\frac{\#\mathrm{training\ samples}}{500}$, which means that each device $c\!\in\!C$ will receive $v_c$ data samples one by one in each communication round, and the samples would be shuffled and appear again per $500$ rounds. 
Other default configurations are shown in Table \ref{tab: statistical information of tasks and datasets}. 
Note that the participating clients in each round are randomly selected, and for each experiment, we repeat $5$ times and show the average results.

\subsection{Baselines}
\label{appendix: introduction to baselines}
(1) \textbf{Random sampling methods} including \textit{RS} (Reservoir Sampling~\cite{vitter1985random}) and \textit{FIFO} (First-In-First-Out: storing the latest $|B_c|$ data samples)\footnote{The experiment results of the two random sampling methods are similar, and thus we choose \textit{RS} for random sampling only.}.
    
(2) \textbf{Importance sampling-based methods} including \textit{HighLoss (HL)}, using the loss of each data sample as data value to reflect the informativeness of data \cite{loshchilov2015online, shrivastava2016training, schaul2015prioritized}, and \textit{GradientNorm (GN)}, quantifying the impact of each data sample on model update through its gradient norm \cite{johnson2018training, zhao2015stochastic}.
    
(3) \textbf{Previous data selection methods} for canonical FL including \textit{FedBalancer (FB)}~\cite{DBLP:conf/mobisys/ShinLLL22} and \textit{SLD} (Sample-Level Data selection)~\cite{li2021sample} which are revised slightly to adapt to streaming data setting: (i) We store the loss/gradient norm of the latest $50$ samples for noise removal; (ii) For \textit{FedBalancer}, we ignore the data samples with loss larger than top $10\%$ loss value, and for \textit{SLD}, we remove the samples with gradient norm larger than the median norm value.
    
(4) \textbf{Ideal case} with unlimited on-device storage, denoted as \textit{FullData (FD)}, using the entire dataset of each client for training to simulate the unlimited storage scenario,.

\subsection{Motivating Experiments}
\label{appendix: motivating experiments}
In this section, We provide the complete experimental evidences for our motivation. First, we prove that the properties of limited on-device storage and streaming data can deteriorate the classic FL model training process significantly in various settings, such as different numbers of local training epochs and different data heterogeneity among clients. Then, we analyze the separate impact of theses two properties on FL with different data selection methods. Due to limited space, we only provide the main conclusions here and the details of the experiment settings and results are provided in the technical report~\cite{technical_report}.

The main results are: 
(1) When the number of local epoch $m$ increases, the negative impact of limited on-device storage is becoming more
serious due to larger steps towards the biased update direction, slowing down the convergence time $3.92\times$ and decreasing the final model accuracy by as high as $6.7\%$;
(2) With the variance of local data increasing, the reduction of convergence rate and model accuracy is becoming larger. This is because the stored data samples are more likely to be biased due to wide data distribution;
(3) The property of streaming data prevents previous methods from making accurate online decisions, as they select each sample according to a normalized probability depending on both discarded and upcoming samples, which are not available in streaming setting. But {\ttfamily ODE} selects each data sample through a deterministic valuation metric, not affected by the other samples;
(4) The property of limited storage is the essential failure of existing data selection methods as it prevents previous methods from obtaining full local and global data information, guiding clients to select suboptimal data from an insufficient candidate dataset. In contrast, {\ttfamily ODE} allows clients to select valuable samples with global information from server.

\subsection{Component-wise Analysis}
\label{appendix: component-wise analysis}
In this subsection, we evaluate the effectiveness of each of the three components in {\ttfamily ODE}: on-device data selection, global gradient estimator and cross-client coordination strategy. The detailed experimental settings, results and analysis are presented in the technical report~\cite{technical_report}, and we only present the main conclusions here. 
    
\textbf{On-Client Data Selection.}
The result shows that without data valuation module, \textit{Valuation-} performs slightly better than \textit{RS} but much worse than \textit{ODE-Est}, which demonstrates the significant role of our data valuation and selection metric.

\textbf{Global Gradient Estimator.}
Experimental results show that using the naive estimation method instead of our proposed local and global gradient estimators will lead to really poor performance, as the partial client participation and biased local gradient will cause the inaccurate estimation for global gradient, which further misleads the clients to select samples using a wrong valuation metric.

\textbf{Cross-Client Coordination.}
Empirical result shows that without the cross-client coordination component, the performance of {\ttfamily ODE} is largely weakened, as the clients tend to store similar and overlapped valuable data samples and the other data will be under-represented.

The results altogether show that each component is critical for the good performance  of {\ttfamily ODE}.

\section{Proofs}
\label{appendix: proofs}
The full proofs of theorems and lemmas are also provided in the technical report~\cite{technical_report} due to limited space.

\end{document}